\documentclass[10pt,twocolumn,letterpaper]{article}

\usepackage{style}
\usepackage{times}
\usepackage{epsfig}
\usepackage{graphicx}
\usepackage{amsmath}
\usepackage{amssymb}

\usepackage{booktabs}

\usepackage[breaklinks=true,bookmarks=false]{hyperref}

\iccvfinalcopy

\ificcvfinal\fi

\begin{document}

\title{In Defense of the Learning Without Forgetting for Task Incremental Learning}

\author{Guy Oren and Lior Wolf\\
Tel-Aviv University\\
{\tt\small \{guyoren347, liorwolf\}@gmail.com}
}

\maketitle
\ificcvfinal\thispagestyle{empty}\fi
\begin{abstract}
   Catastrophic forgetting is one of the major challenges on the road for continual learning systems, which are presented with an on-line stream of tasks. The field has attracted considerable interest and a diverse set of methods have been presented for overcoming this challenge. Learning without Forgetting (LwF) is one of the earliest and most frequently cited methods. It has the advantages of not requiring the storage of samples from the previous tasks, of implementation simplicity, and of being well-grounded by relying on knowledge distillation. However, the prevailing view is that while it shows a relatively small amount of forgetting when only two tasks are introduced, it fails to scale to long sequences of tasks. This paper challenges this view, by showing that using the right architecture along with a standard set of augmentations, the results obtained by LwF surpass the latest algorithms for task incremental scenario. This improved performance is demonstrated by an extensive set of experiments over CIFAR-100 and Tiny-ImageNet, where it is also shown that other methods cannot benefit as much from similar improvements. Our code is available at: \url{https://github.com/guy-oren/In_defence_of_LWF}
\end{abstract}

\section{Introduction}

The phenomenon of catastrophic forgetting (CF) of old concepts as new ones are learned in an online manner is well-known. The approaches to overcome it can be categorized, as suggested by De Lange \etal \cite{de2019continual}, into three families: (i) replay-based methods, which store selected samples of previously encountered classes, (ii) regularization-based methods, that limit the freedom to learn new concepts, and (iii) parameter isolation methods, which directly protect the knowledge gained in the past, by dividing the network parameters into separate compartments.

The field of continual learning is very active, with dozens of methods that have emerged in the last few years.  However, it seems that the growing interest leads to confusion rather than to the consolidation of knowledge. As practitioners looking to find out which online learning method would be suitable for a real-world application, we were unable to identify the solid methods of the field and could not infer from the literature the guiding principles for tackling catastrophic forgetting.

Indeed, reviewing the literature, one can find many insightful ideas and well-motivated solutions. However, little data regarding the generality of continual learning methods, the sensitivity of the methods to the specific setting and hyperparameters, the tradeoff between memory, run-time and performance, and so on. Ideally, one would like to find a method that is not only well-grounded and motivated, but also displays a set of desired properties: (i) work across multiple datasets, (ii) be stable to long sequences of on-line learning tasks, (iii) benefit from additional capacity, (iv) display flexibility in network architecture that allows the incorporation of modern architectures, (v) display an intuitive behavior when applying regularization, and (vi) present robustness to hyperparameters.

We demonstrate that these properties hold for one of the first methods to be proposed for tackling {CF}, namely the Learning without Forgetting (LwF) method~\cite{li2017learning}. This is a bit surprising, since this method, as a classical method in a fast-evolving field, has been repeatedly used as an inferior baseline. However, we show that unlike many of the more recent methods, this scapegoat method can benefit from residual architectures and further benefits from simple augmentation techniques. Moreover, while the original LwF implementation employed techniques such as warmup and weight decay, we were able to train without these techniques and their associated hyperparameters. Overall, we find LwF, which is a simple data-driven regularization technique, to be more effective than the most promising regularization-based and parameter-isolation methods.

\section{Related work}

It is often the case that new methods are presented as having clear advantages over existing ones, based on empirical evidence. The inventors of these methods have little incentive to explore the underlying reason for the performance gap. Without a dedicated effort to do so, the literature can quickly become misleading. 

In our work, we demonstrate that the task-incremental learning methods that have emerged since the 2016 inception of the LwF method are not more accurate than this straightforward method. This demonstration is based on changing the underlying neural network architecture to a ResNet~\cite{he2016deep} and on employing a simple augmentation technique during training. Moreover, we show that LwF benefits from more capacity, width wise.

A recent related attempt by De Lange \etal \cite{de2019continual} also addresses the need to compare multiple continual learning algorithms in task-incremental settings. That study has employed multiple architectures, and, similar to us, have noted that the LwF method benefits from the additional capacity given by extra width but not from extra depth. However, ResNets or augmentations were not employed and the conclusion was that LwF is not competitive with the more recent techniques.  This conclusion is in sheer contrast to ours, demonstrating the challenge of comparing methods in a way that exposes their full potential, and the need to perform such comparative work repeatedly.

\subsection{Task-incremental learning}

CF in neural networks has been observed from the beginning. However, there is no consensus regarding the proper settings and metrics for comparing different techniques. In this work, we adopt a setting definition from the work of \cite{van2019three, Hsu2018ReevaluatingCL}, who define three different settings for continual learning -- task incremental, domain incremental, and class incremental. In all scenarios, the system is presented with a stream of tasks and is required to solve all tasks that are seen so far. In task incremental, the task identifier is provided both in train and inference time. In domain incremental, the task identifier is provided only in train time, and the classifier does not need to infer the task identifier but rather just solve the task at hand. In class incremental, the learner also needs to infer the task identifier in inference time. 

We focus on the task incremental setting. Moreover, we do not consider replay-based methods since these rely heavily on accessing data retained from the previous tasks, which is not desirable in real-world scenarios, and depends on an additional parameter that is the size of the memory.

The literature has a great number of methods, further emphasizing the need for comparative work. In this work, we focus on the methods that are repeatedly reported in the literature~\cite{de2019continual,serra2018overcoming,Hu2019,Li2019}. These include: Elastic Weight Consolidation (EWC; \cite{kirkpatrick2017overcoming}, online version), Incremental Moment Matching (IMM; \cite{lee2017overcoming}, both Mean and Mode variants), overcoming CF with Hard Attention to the Task (HAT; \cite{serra2018overcoming}), continual learning with Hypernetworks (Hyper-CL; \cite{von2019continual}) and Adversarial Continual Learning (ACL; \cite{Ebrahimi2020AdversarialCL}).

Both the EWC and IMM variants, belong to a regularization-based family and add a structural, weight-based, regularization term to the loss function to discourage changes to weights that are important for previous tasks. IMM performs a separate model-merging step after learning a new task, which EWC does not. Although this family of methods is very rich, IMM and EWC are among the leading methods and are often cited as baselines. 

The HAT approach belongs to the parameter isolation family and applies a light-weight, unit-based, learnable, and 'soft' masks per task. HAT is a successor to various works, including (i) progressive neural networks (PNNs; \cite{rusu2016progressive}), which applies a complete and separate network for each task (columns) with adapters between columns, (ii) PathNet \cite{fernando2017pathnet} that also pre-assigns some amount of network capacity per task but, in contrast to PNNs, avoids network columns and adapters and instead suggests to learn evolutionary the paths between modules, and (iii) PackNet \cite{mallya2018packnet}, which uses weight-based pruning heuristics and a retraining phase to maintain a binary mask for each task. Since HAT was shown to have both performance and computational advantages over (i)-(iii), we focus on it as a representative method from this line of work.

Hyper-CL~\cite{von2019continual}, a recent addition to the parameter isolation family, belongs to a different branch in this family than HAT. Instead of using a fixed pre-determined capacity, Hyper-CL suggests learning the weights of a target network for each task. Hyper-CL employs a variant of Hypernetworks \cite{ha2016hypernetworks}, called Chunked-Hypernetworks~\cite{pawlowski2017implicit}, which generates different subsets of the target network's parameters using the same generator. To do so, the method learns both the task embedding and the ``chunk'' embedding. This variant makes it possible to maintain a much smaller hypernetwork than the target network. To overcome CF, they apply regularization that constrains the weights of the previously seen target task {from changing}.

Some methods belong to more than one category. ACL~\cite{Ebrahimi2020AdversarialCL} employs both parameter isolation using a small private network for each task, and regularization for a shared network across tasks. This regularization contains two parts: an adversarial loss that makes the shared encoding task-independent~\cite{Ganin2016DomainAdversarialTO} and a disentanglement loss that acts to remove the overlap between the private- and the shared-encoding ~\cite{Salzmann2010FactorizedOL}.

Naturally, given the number of relevant methods, it is not feasible to compare with all of them. The regularization-based family presents two additional methods that we considered: Encoder Based Lifelong Learning (EBLL; \cite{rannen2017encoder}) and Memory Aware Synapses (MAS; \cite{aljundi2018memory}). EBLL extends LwF by adding a per-task auto-encoder, requiring further hyperparameter tuning. The literature shows that it only marginally improves over LwF for AlexNet-like architectures~\cite{de2019continual,aljundi2018memory}, and our attempts to apply it together with ResNets led to poor results. MAS was also shown in \cite{de2019continual} to only slightly improved over LwF.

\section{The LwF method and its modifications}

The LwF method by Li \etal \cite{li2017learning}, belongs to the regularization-based family. However, unlike EWC and IMM, its regularization is data-driven. The method seeks to utilize the knowledge distillation loss~\cite{hinton2015distilling} between the previous model and the current model to preserve the outputs of the previous task. Since maintaining the data of previous tasks is not desirable and rather not scalable, LwF uses only the current task data for knowledge distillation.

In the task-incremental setting, the learner is given a new set of labels to learn at each round. This set of classes is called a task. In LwF the classifier is composed out of two parts: the feature extractor $f$ and a classifier head $c_i$ per each task for $i=1,2, \dots, T$.

Let $\{(x_j^t,y_j^t)\}$ be the set of training samples for task t. The cross-entropy loss is used as the primary loss for training the classifier $c_t \circ f$:
\begin{equation}
    L_{CE} = -\sum_j \log [c_t(f(x^t_j))]_{y_j^t}
\end{equation}
where the subscript $y_j^t$ is used to denote the pseudo-probability of the classifier for the ground truth label. 

When learning a new task $t$, to maintain previous task knowledge, we employ knowledge distillation between the ``old'' feature extraction and the previous task classifier heads and the new ones. These are denoted by $f^o$ for the previous feature extractor network (as learned after task $t-1$), and $c_i^o$ for $i=1,2, \dots ,t-1$ for the previous heads. The learned feature extraction is denoted by $f$ and the updated task classifiers are denoted by $c_i$, for $i=1,2, \dots t$. 

For simplicity, we described the knowledge distillation process for one previous task and one sample $(x,y) \in \{(x_j^t,y_j^t)\}$ from the current task $t$. However, the process is repeated for the classifier heads of all previous tasks and all samples of task $t$, while summing up the individual losses. Let $Y^o := [y^o_1,y^o_2,...] = c^o_i(f^o(x))$ be the vector of probabilities that the old classifier of task $i$ assigns to sample $x$. Similarly, let $Y:=[y_1,y_2,...]$ be the vector of probabilities for the same training samples obtained with $c_i \circ f$. To apply the knowledge distillation loss, these vectors are modified in accordance with some temperature parameter $\theta$: 

\begin{align}
    y'_k = \frac{y_k^{\frac{1}{\theta}}}{\sum_{m}y_m^{\frac{1}{\theta}}}\, ,\quad
    {y_k'}^{o} = \frac{(y_k^{o})^{\frac{1}{\theta}}}{\sum_{m}(y^o_m)^{\frac{1}{\theta}}}\, .
\end{align}
The temperature is taken to be larger than one, to increase small probability values and reduce the dominance of the high values. The knowledge distillation loss is defined as:
\begin{align}
    L_{dist} = - \sum_{k}{y'_k}^o log(y'_k)\,,
\end{align}
\noindent where the summation is done over all labels of task $i$.

We followed the author's suggestions and in all our experiments and set $\theta = 2$ and the regularization weight to one, \ie, the knowledge distillation loss had the same weight as the classification loss of the new task. It is worth mentioning that although the original LwF work~\cite{li2017learning} evaluated the method in the two task scenario, it can be readily extended to any number of tasks by using knowledge distillation loss over all $c^o_i , i=1,2 \dots, t-1$. This further highlights the need for performing our research, since such an extension was previously done in the context of attempting to present the preferable performance of a new method. We also note that it was suggested in~\cite{li2017learning} to use a warmup phase at the beginning of training for each new task, in which both $f$ and $c_i, i=1,2,\dots, t-1$ are frozen and one trains $c_t$ with the cross-entropy loss until convergence. However, since the effect of this seems negligible even in the original paper, we do not perform this. The authors also used regularization in the form of weight decay during training, which we remove to avoid the need to fit a regularization hyperparameter for each experiment. Moreover, in our initial experiments weight decay tends to hurt the accuracy of new tasks.

\subsection{Architecture}

 Li \etal \cite{li2017learning} employed AlexNet~\cite{krizhevsky2012imagenet} and VGGNet~\cite{simonyan2014very} to evaluate the performance of the method. Interestingly, even the recent review work by De Lange \etal \cite{de2019continual} uses AlexNet as a reference network, despite ongoing advances in network architectures. {There is also a key difference between the different versions of AlexNet-like architectures employed in \cite{li2017learning} and \cite{serra2018overcoming}. The latter use Dropout \cite{srivastava2014dropout}, which as we show empirically, is detrimental}. 
 
 We also offer to use the ResNet~\cite{he2016deep} architecture.  We are not the first to attempt to use ResNets for LwF. Mallya \etal \cite{mallya2018packnet} employed LwF with a ResNet-50 network as an underperforming baseline. { However, our} experiments demonstrate that LwF mostly benefits from a Wide-ResNet~\cite{zagoruyko2016wide} network rather than from deeper ones.

\subsection{Data augmentation}
\label{sec:data_aug}

Using a method with a shared model presents a challenge. On the one hand, the shared part must have enough capacity to learn new tasks. On the other hand, bigger networks are more vulnerable to overfitting when training on the first tasks. The parameter isolation family works around this problem by dynamically changing the capacity of the network as in PNNs~\cite{rusu2016progressive} or learning a specific target network for each task with enough capacity for each task, like in Hyper-CL \cite{von2019continual}.

In addition to the capacity needs, another challenge that the LwF method faces is the need to mitigate the difference between the input distributions for different tasks. In the extreme, where the input distributions are very dissimilar, the knowledge distillation loss is no longer constraining the network to success on previous tasks. 

Data augmentation, which is a well-studied technique for overcoming overfitting by virtually expending the dataset at hand, also has the potential to close the gap between different input distributions and therefore reduce forgetting. In our experiments, we employ a very basic set of augmentation consisting of random horizontal flips, color jitter (randomly change the brightness, contrast, saturation, and hue), and translation. As it turns out, these are sufficient to reduce the forgetting almost to zero, while substantially increasing the average accuracy for all tested settings.

\section{Experiments}

The common datasets for evaluating CF in classification problems include permutations of the MNIST data~\cite{srivastava2013compete}, a split of the MNIST data~\cite{lee2017overcoming}, incrementally learning classes of the CIFAR data sets~\cite{lopez2017gradient}, or on considering two datasets and learning the transfer between them~\cite{li2017learning}. Serr{\`a} \etal \cite{serra2018overcoming} points out the limitations of the MNIST setups, since these do not well represent modern classification tasks. The two-task scenario is criticized for being limited and does not enable the evaluation of CF for sequential learning with more than two tasks. CIFAR-100 splits are criticized for having tasks that are relatively similar in nature. However, in our experiments, performance on CIFAR-100 {splits} discriminates well between different methods and between different settings of the same method.

In addition to CIFAR-100~\cite{krizhevsky2009learning}, we employ Tiny-ImageNet~\cite{Le2015TinyIV} in our experiments. 
The latter presents a higher diversity with more classes and the ability to challenge methods with longer and more meaningful sequences of tasks. To obtain a generic estimate, we shuffle the order of classes in each dataset and repeat each experiment setup five times with different seeds.

A common CIFAR setup, introduced in \cite{zenke2017continual} offers to use CIFAR-10 as a first task, then split CIFAR-100 into five distinct tasks with 10 disjoint classes each. However, it may introduce a bias in evaluating task-incremental methods, since it makes the first task much larger and, therefore, conceals the problem of first-task overfitting. In this work, we consider a different setting, in which CIFAR-100 is divided into 5-Splits (i.e., 5-tasks), 10-Splits, and 20-Splits with 20, 10, and 5 classes in each task, respectively. Each class in CIFAR-100 contains 500 training images and 100 testing images. Each image size is $3 \times 32 \times 32$. As a validation set, we shuffle the training data and use $90\%$ as training examples and $10\%$ as validation examples.

A recent work by De Lange \etal \cite{de2019continual} employed Tiny-ImageNet as a benchmark using a  similar setup to the CIFAR-100 setup above. However, they split the dataset to 20 disjoint tasks with 10 classes each. Since we opt for a longer sequence of tasks while still keeping them meaningful, we split the dataset into 40 disjoint tasks with 5 classes each. As our results will show, this setting pushes the limits of the task-incremental methods.

Each class in Tiny-ImageNet contains 500 training images, 50 validation images, and 50 testing images. The original image size for this dataset is $3 \times 64 \times 64$. 
Since the test set is not publicly available, we use the validation set as a test set and as a validation set, we shuffle the training data and use $90\%$ for training and $10\%$ for validation.

To evaluate performance, we adopt the metrics of \cite{lopez2017gradient}:
{\small\begin{align}
  \textbf{Average Accuracy:~~} \text{ACC} &= \frac{1}{T}\sum_{i=1}^{T}R_{T,i} \label{eq:acc}\\
  \textbf{Backward Transfer:} \text{ BWT} &= \frac{1}{T - 1}\sum_{i=1}^{T - 1}R_{T,i} - R_{i, i} \label{eq:bwt}
\end{align}}
where $T$ is the number of tasks and $R_{i,j}$ is the test accuracy score for task $j$ after the model learned task $i$. We note that $BWT < 0$ reports CF, while $BWT > 0$ indicates that learning new tasks helped the preceding tasks.

\subsection{The effect of the network architecture}
\label{sec:net_effect}
We first present experiments for LwF with various network architectures and no data augmentation. The AlexNet-like architecture~\cite{krizhevsky2012imagenet} we use follows~\cite{serra2018overcoming} and has three convolutional layers of 64, 128, and 256 filters with $4 \times 4$, $3 \times 3$, and $2 \times 2$ kernel sizes, respectively. On top, there are two fully-connected layers of 2048 units each. This network employs rectified linear units (ReLU) as activations, and $2 \times 2$ max-pooling after the convolutional layers. A Dropout of 0.2 is applied for the first two layers and 0.5 for the rest. All layers are randomly initialized with Xavier uniform initialization \cite{glorot2010understanding}.

\begin{table*}[t]
\centering
\resizebox{\textwidth}{!}{
\begin{tabular}[t]{@{}l@{~}l@{~}c@{~}c@{~~}c@{~}c@{~~}c@{~}c@{~~}c@{~}c@{}}
\toprule
& & \multicolumn{2}{c}{CIFAR 5-Split}     & \multicolumn{2}{c}{CIFAR 10-Split}    & \multicolumn{2}{c}{CIFAR 20-Split}  & \multicolumn{2}{c}{Tiny-ImageNet 40-Split}    \\
\cmidrule(l{2pt}r{2pt}){3-4}
\cmidrule(l{2pt}r{1pt}){5-6}
\cmidrule(l{2pt}r{1pt}){7-8}
\cmidrule(l{2pt}r{1pt}){9-10}
Arch.    & \#Params & BWT            & ACC            & BWT            & ACC            & BWT            & ACC             & BWT            & ACC             \\
\midrule
AlexNet-D  & $6.50$ &$-39.9 \pm 1.4$  & $36.6 \pm 1.5$    & $-52.9 \pm 1.2$   & $28.1 \pm 1.3$   & $-54.4 \pm 1.1$   & $31.3 \pm 0.8$    & $-50.5 \pm 1.0$  & $25.0 \pm 0.4$   \\
AlexNet-ND  & $6.50 $ & $-1.8 \pm 0.6$  & $56.6 \pm 1.1$    & $-2.9 \pm 0.2$   & $67.0 \pm 1.0$   & $-3.1 \pm 0.3$   & $75.5 \pm 0.6$    & $-2.8 \pm 0.3$  & $66.9 \pm 0.8$ \\
RN-20    & $0.27$ & $-0.4 \pm 0.3$ & $60.4 \pm 0.7$ &  $-1.9 \pm 0.5$ &  $67.2 \pm 1.0$  &  $-2.3 \pm 0.4$ &  $76.2 \pm 0.8$  & $-3.0 \pm 0.5$  & $70.8 \pm 1.0$   \\
RN-32  & $0.47$ &  $-1.8 \pm 0.7$ & $58.8 \pm 2.0$  &  $-1.8 \pm 0.2$ & $67.1 \pm 1.1$   &   $-2.7 \pm 0.2$ &  $75.6 \pm 0.4$ & $-2.4 \pm 0.2$  &  $70.9 \pm 1.1$  \\
RN-62  & $0.95$ & $-1.7 \pm 0.6$  & $58.9 \pm 0.7$  &  $-2.7 \pm 0.4$ &  $66.0 \pm 0.8$  &  $-2.9 \pm 0.4$ & $75.6 \pm 0.7$  &  $-3.1 \pm 0.9$ & $70.3 \pm 1.2$   \\
WRN-20-W2 & $1.08$ & $-1.2 \pm 0.6$ & $62.0 \pm 0.3$  & $-2.1 \pm 0.6$   & $69.6 \pm 0.8$ & $-3.3 \pm 0.4$ & $77.3 \pm 0.4$ &  $-3.8 \pm 0.2$  &  $71.5 \pm 0.6$ \\
WRN-20-W5 & $6.71$ & $-2.0 \pm 0.5$ & $64.2 \pm 1.1$  & $-2.9 \pm 0.3$   & $71.2 \pm 0.5$ & $-3.7 \pm 0.3$ & $79.4 \pm 0.6$ & $-4.5 \pm 0.3$   & $72.6 \pm 0.8$  \\
\end{tabular}
}
\caption{\small Network results summary for LwF. BWT and ACC in \%. \#Params in millions and counts only for the shared feature extractor. All results are averaged over five runs with standard deviations. {D=Dropout, ND=No Dropout, }RN=ResNet, WRN=WideResNet.}
\label{table:arch_table}
\end{table*}

{While LwF is commonly used with an AlexNet-like architecture \cite{Li2019,serra2018overcoming,de2019continual}, we opt to use more modern architectures.} We choose to use the popular architecture family of ResNets. In this work, we use ResNet-20 (RN-20), ResNet-32 (RN-32) and ResNet-62 (RN-62)~\cite{he2016deep}, as well as Wide-ResNet-20 networks with width factors 2 or 5~\cite{zagoruyko2016wide} (WRN-20-W2 and WRN-20-W5 respectively). Those networks employ ReLU activations and Batch Normalization layers~\cite{ioffe2015batch}. All convolutional layers were randomly initialized with Kaiming normal inits with fan-out mode~\cite{he2015delving}, and the normalization layers were initialized as constants with 1 and 0 for weight and bias, respectively. All architecture tested use separated fully-connected layers with a softmax output for each task as a final layer. More details can be found in the appendix.

In all experiments, LwF is trained up to 200 epochs for each task. We use a batch size of 64 and an SGD optimizer with a learning rate of $0.01$ and a momentum of $0.9$. We used the validation set to schedule the learning rate, where we drop the learning rate by a factor of 3 if there is no improvement in the validation loss for five consecutive epochs. Training is stopped when the learning rate becomes lower than $10^{-4}$. 

The results are depicted in Tab.~\ref{table:arch_table}. Our clearest and most significant result is that the underlying network has a great effect on LwF performance. While LwF with AlexNet {with Dropout} architecture greatly suffers from forgetting which results in low ACC, just removing the Dropout from the network results in a sizable performance boost. This makes sense while using Dropout on the teacher side creates a strong teacher that can be viewed as a large ensemble of models that shares weight \cite{hinton2015distilling}, on the student side, this weakens the regularization of LwF. Randomly choosing which weights to regularize ignores their importance for older tasks, which results in high forgetting. 

Next, switching to RN-20 with an order of magnitude fewer parameters shows preferable performance. This change reveals the potential of LwF to obtain competitive ACC and BWT.

Following \cite{de2019continual} we investigate the effect of width and depth {of the architecture} with the ResNet network on LwF performance. We used two deeper networks (RN-32 and RN-62) and two wider networks (WRN-20-W2 and WRN-20-W5). Our results (Tab.~\ref{table:arch_table}) show that while using a deeper network gives similar or inferior results compare to RN-20, using wider networks increases performance. 

\subsection{The effect of data augmentation}

We conjectured in Sec.~\ref{sec:data_aug} that LwF performance can be further increased by using data augmentations. In this section, we conduct experiments on WRN-20-W5, which is the best performer among the tested architectures, with a relatively simple set of random augmentations:  random horizontal translation of up to 3 pixels with reflection padding, random horizontal flip, and color jitter (brightness, contrast and saturation with jitter of $0.3$ and hue with jitter of $0.2$). 

\begin{table*}[t]
\resizebox{\textwidth}{!}{
\centering
\begin{tabular}[t]{@{}l@{~~}c@{~}c@{~~}c@{~}c@{~~}c@{~}c@{~~}c@{~}c@{}}
\toprule
& \multicolumn{2}{c}{CIFAR 5-Split}     & \multicolumn{2}{c}{CIFAR 10-Split}    & \multicolumn{2}{c}{CIFAR 20-Split}  & \multicolumn{2}{c}{Tiny-ImageNet 40-Split}    \\
\cmidrule(l{2pt}r{2pt}){2-3}
\cmidrule(l{2pt}r{1pt}){4-5}
\cmidrule(l{2pt}r{1pt}){6-7}
\cmidrule(l{2pt}r{1pt}){8-9}
Augmentation    & BWT            & ACC            & BWT            & ACC            & BWT            & ACC             & BWT            & ACC             \\
\midrule
Without  & $-2.0 \pm 0.5$ & $64.2 \pm 1.1$  & $-2.9 \pm 0.3$   & $71.2 \pm 0.5$ & $-3.7 \pm 0.3$ & $79.4 \pm 0.6$ & $-4.5 \pm 0.3$   &  $72.6 \pm 0.8$ \\
With  & $-0.2 \pm 0.2$ & $80.3 \pm 0.6$  &  $-0.6 \pm 0.2$  & $83.7 \pm 0.8$ & $-1.5 \pm 0.3$ & $86.6 \pm 0.4$ &  $-2.1 \pm 0.2$  & $78.6 \pm 0.6$  \\
    \end{tabular}
}
    \caption{\small Data augmentation results for LwF with WRN-20-W5 architecture. {BWT and ACC in \%. All results are averaged over five runs with standard deviations.} }
    \label{table:aug_table}
\end{table*}
\begin{table*}[t]
\resizebox{\textwidth}{!}{
\centering
\begin{tabular}[t]{@{}l@{~}l@{}c@{~}c@{~}c@{~~}c@{~}c@{~~}c@{~}c@{~~}c@{~}c@{}}
\toprule
& & & \multicolumn{2}{c}{CIFAR 5-Split}     & \multicolumn{2}{c}{CIFAR 10-Split}    & \multicolumn{2}{c}{CIFAR 20-Split}  & \multicolumn{2}{c}{Tiny-ImageNet 40-Split}    \\
\cmidrule(l{2pt}r{2pt}){4-5}
\cmidrule(l{2pt}r{1pt}){6-7}
\cmidrule(l{2pt}r{1pt}){8-9}
\cmidrule(l{2pt}r{1pt}){10-11}
Method & Arch. & Aug.    & BWT            & ACC            & BWT            & ACC            & BWT            & ACC             & BWT            & ACC             \\
\midrule
EWC & AlexNet-D & & $+0.2 \pm 0.1$ & $58.6 \pm 0.9$  & $+0.7 \pm 0.4$  &  $64.1 \pm 0.5$  & $+0.0 \pm 0.9$  & $74.0 \pm 1.0$ & $-0.8 \pm 0.4$ & $63.3 \pm 0.9$   \\
EWC & AlexNet-D & \checkmark & $+0.0 \pm 0.2$ & $62.9 \pm 1.5$ &  $+0.1 \pm 0.4$  & $68.4 \pm 0.9$ & $-0.5 \pm 1.1$ & $75.2 \pm 1.3$   & $-1.5 \pm 2.0$ & $63.8 \pm 2.6$ \\
IMM-MEAN & AlexNet-D &  & $-1.2 \pm 0.8$ & $58.9 \pm 1.1$  & $-0.6 \pm 0.7$ & $58.6 \pm 1.9$  & $-0.8 \pm 0.3$ & $55.9 \pm 1.6$ &  $-0.6 \pm 0.8$  & $43.6 \pm 1.3$  \\
IMM-MEAN & AlexNet-D & \checkmark & $-2.5 \pm 1.0$ & $62.5 \pm 1.8$  & $-1.3 \pm 0.8$ & $61.4 \pm 2.0$  & $-1.3 \pm 0.5$ & $57.9 \pm 2.9$ &  $-1.2 \pm 0.5$  &  $44.7 \pm 1.5$ \\
IMM-MODE & AlexNet-D & & $-8.3 \pm 1.5$ & $63.7 \pm 1.5$ &  $-21.7 \pm 2.9$  & $58.6 \pm 2.9$  & $-30.5 \pm 3.2$ & $54.9 \pm 3.0$  &  $-25.0 \pm 1.4$  & $50.6 \pm 1.7$   \\
IMM-MODE & AlexNet-D & \checkmark  & $-6.9 \pm 0.3$ & $68.9 \pm 0.9$ & $-19.8 \pm 2.7$ &  $64.4 \pm 2.9$ & $-31.1 \pm 4.2$ & $58.2 \pm 4.3$  &  $-24.2 \pm 2.4$  & $54.6 \pm 2.9$  \\
HAT & AlexNet-D &  & $+0.0 \pm 0.0$   & $67.1 \pm 0.6$   & $+0.0 \pm 0.0$   & $72.8 \pm 0.8$   & $+0.0 \pm 0.0$    & $76.6 \pm 0.6$   & $+0.0\pm 0.0$   & $65.9 \pm 1.1$   \\
HAT & AlexNet-D & \checkmark & $-0.1 \pm 0.0$   & $70.5 \pm 0.9$  & $+0.0 \pm 0.0$  & $76.2 \pm 0.8$  & $+0.0 \pm 0.0$   & $78.4 \pm 1.0$  & $+0.0 \pm 0.0$   & $67.3 \pm 0.9$   \\
HyperCL & H:Lin,M:RN32 &  & $+0.0 \pm 0.1$  & $53.0 \pm 2.3$  &  $+0.0 \pm 0.0$  & $62.9 \pm 0.4$  & $+0.0 \pm 0.0$ & $75.5 \pm 1.0$ & $-0.8 \pm 0.3$   &  $48.9 \pm 1.6$ \\
HyperCL & H:Lin,M:RN32 & \checkmark & $+0.0 \pm 0.0$   & $69.5 \pm 1.1$   & $+0.0 \pm 0.0$  & $78.2 \pm 0.6$   & $+0.0 \pm 0.0$    & $85.3 \pm 0.9$   & $-0.9 \pm 0.3$ & $60.7 \pm 0.3$ \\
$\text{ACL}^o$ & $\text{AlexNet-D}^{**}$ &  & - & - &  - & - & $+0.0 \pm 0.0$ & $78.0 \pm 1.2$   & - & - \\
LwF & WRN-20-W5 & & $-2.0 \pm 0.5$ & $64.2 \pm 1.1$  & $-2.9 \pm 0.3$   & $71.2 \pm 0.5$ & $-3.7 \pm 0.3$ & $79.4 \pm 0.6$ & $-4.5 \pm 0.3$   &  $72.6 \pm 0.8$ \\
LwF & WRN-20-W5 & \checkmark & $-0.2 \pm 0.2$ & $80.3 \pm 0.6$  &  $-0.6 \pm 0.2$  & $83.7 \pm 0.8$ & $-1.5 \pm 0.3$ & $86.6 \pm 0.4$ & $-2.1 \pm 0.2$  & $78.6 \pm 0.6$  \\
\midrule
$\text{JOINT}^*$ & WRN-20-W5 & & $+4.5 \pm 2.0$ & $72.3 \pm 1.9$ & $+4.2 \pm 1.9$  & $80.2 \pm 2.0$   & $+3.0 \pm 1.1$ & $86.1 \pm 0.9$ & $+3.5 \pm 0.3$ & $80.3 \pm 0.3$ \\
$\text{JOINT}^*$ & WRN-20-W5 & \checkmark & $+2.4 \pm 0.8$ &  $85.3 \pm 0.5$ &  $+2.3 \pm 0.2$  & $89.9 \pm 0.4$ & $+1.7 \pm 0.6$ & $93.2 \pm 0.4$ &  $+2.2 \pm 0.5$  & $86.7 \pm 0.4$  \\
    \end{tabular}}
    \caption{\small Comparison between multiple methods. BWT and ACC in \%. *JOINT does not adhere to the task incremental setup, and is performed in order to serve as the upper bound for LwF. **Slightly different AlexNet-like architecture than used in HAT with a similar capacity. $^o$results reported in \cite{Ebrahimi2020AdversarialCL}; all other results are reproduced by us and are averaged over five runs with standard deviations. D=Dropout, RN=ResNet, WRN=WideResNet, Lin=a linear layer, H=Hypernetwork, M=Target network. }
    \label{table:method_table}
\end{table*}

\begin{figure*}[t]
  \centering
    \begin{tabular}{@{}c@{~}c@{~}c@{~}c@{}}
  \includegraphics[width=0.24840924\textwidth]{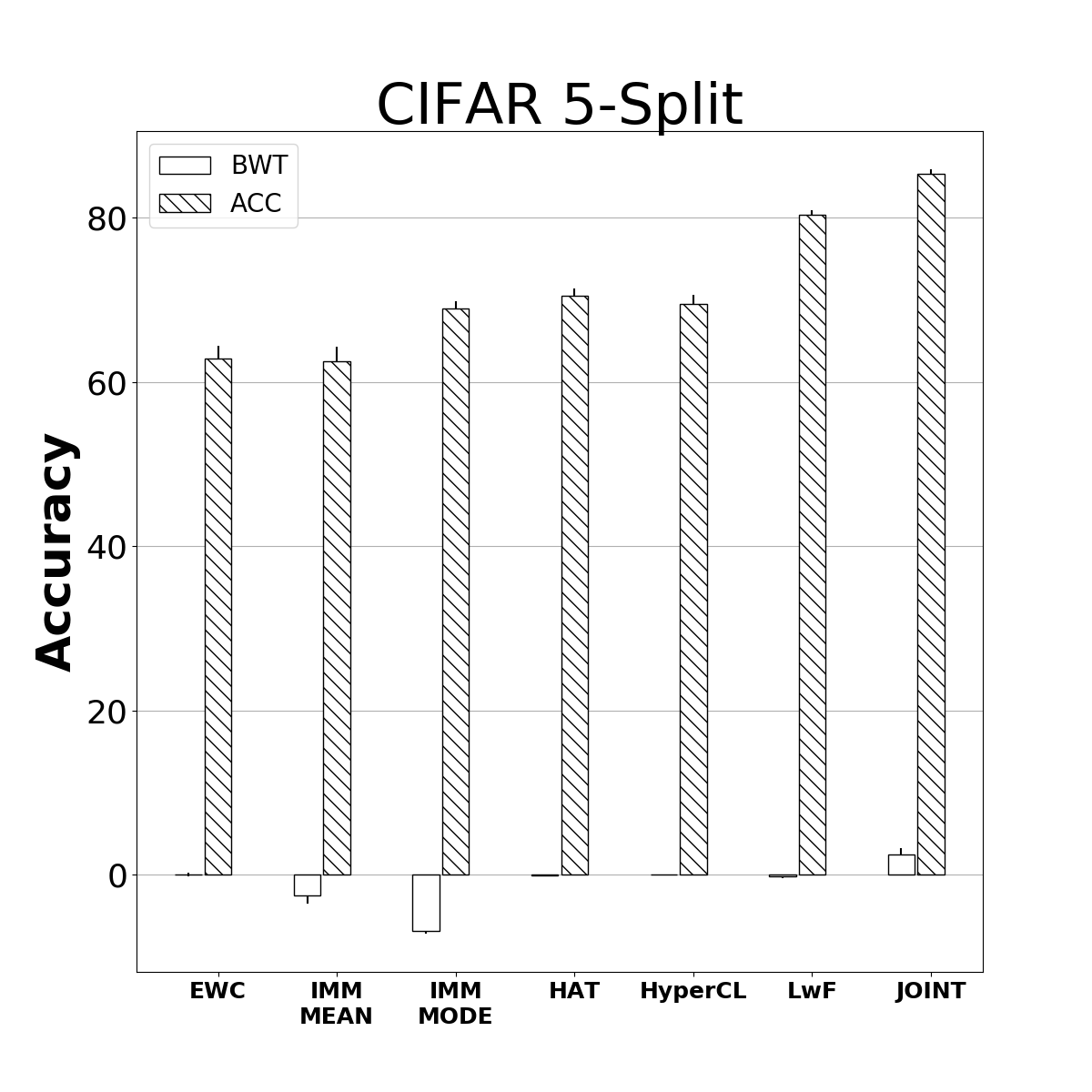} & \includegraphics[width=0.24840924\textwidth]{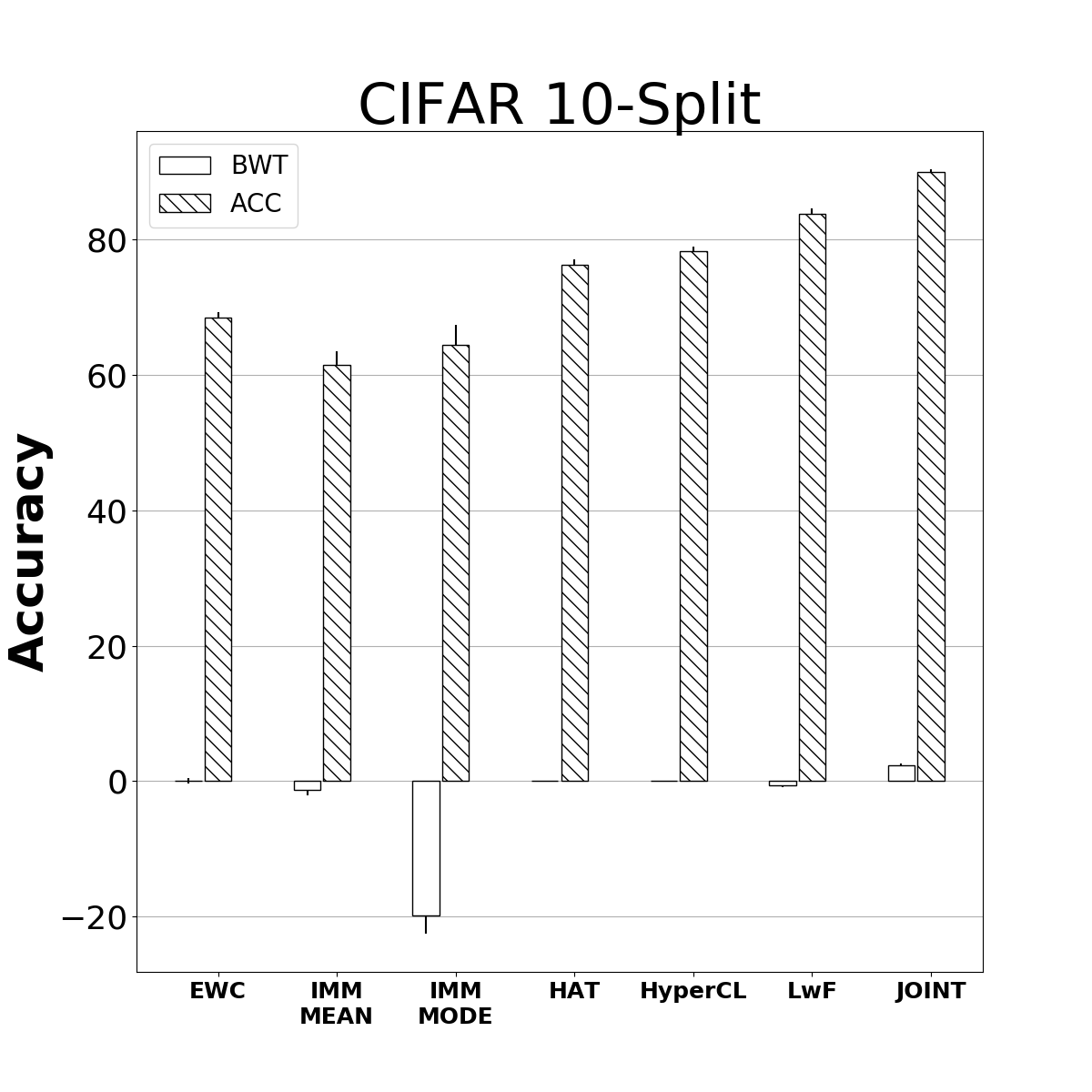} &
  \includegraphics[width=0.24840924\textwidth]{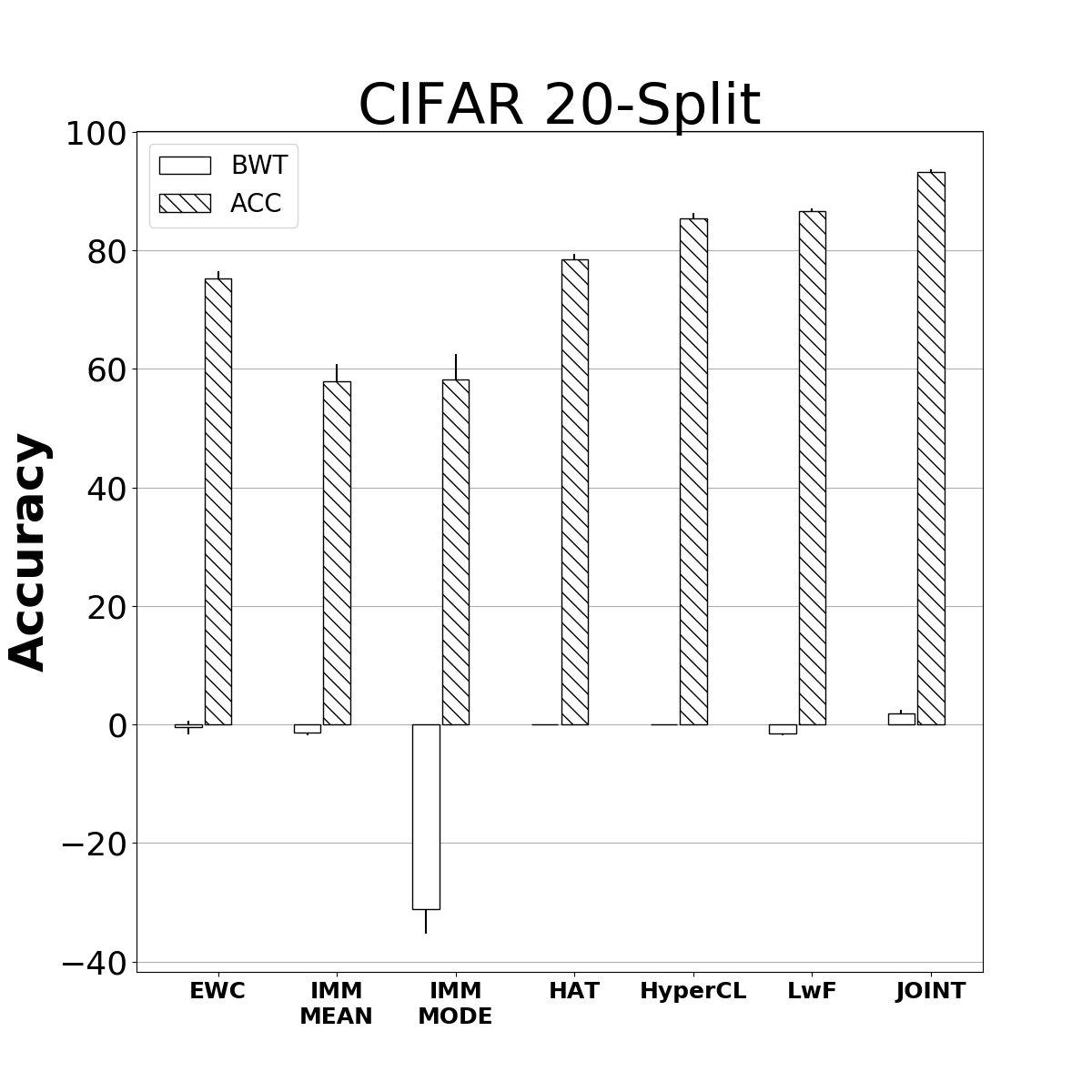} &
  \includegraphics[width=0.24840924\textwidth]{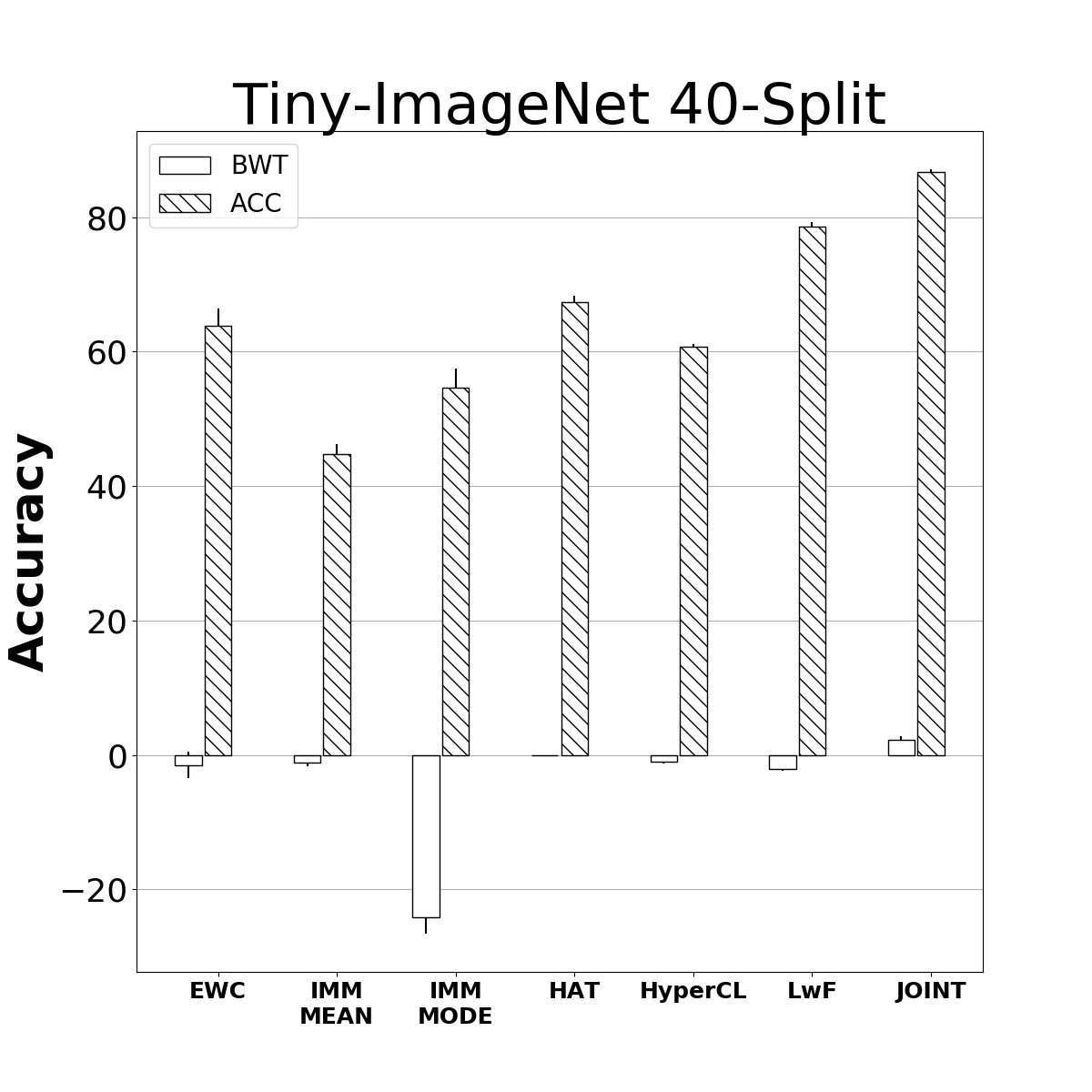}\\
(a) & (b) & (c) & (d)\\
  \end{tabular}
  \smallskip
  \caption{$BWT$ and $ACC$ of the best performance obtained for each of the evaluated methods average over 5 random seeds.
  JOINT is an upper-bound training on all past tasks data. (a) CIFAR 5-Split, (b) CIFAR 10-Split, (c) CIFAR 20-Split, (d) Tiny-ImageNet 40-Split.}
\label{fig:best}
\medskip
  \centering
   \begin{tabular}{cc}
  \includegraphics[width=0.42682436402\textwidth]{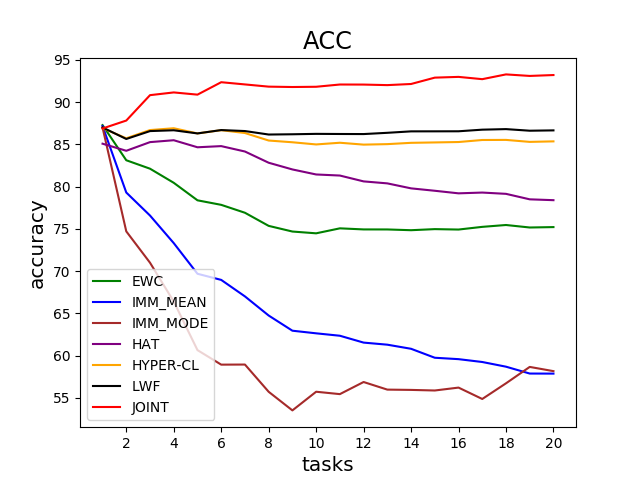} & \includegraphics[width=0.42682436402\textwidth]{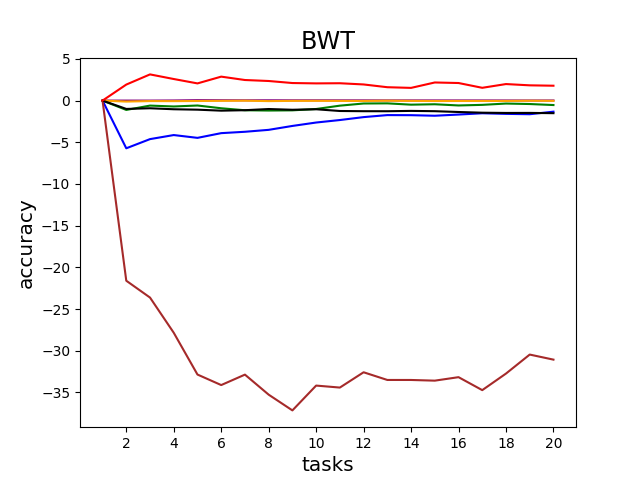} \\
    (a) & (b)\\
  \includegraphics[width=0.42682436402\textwidth]{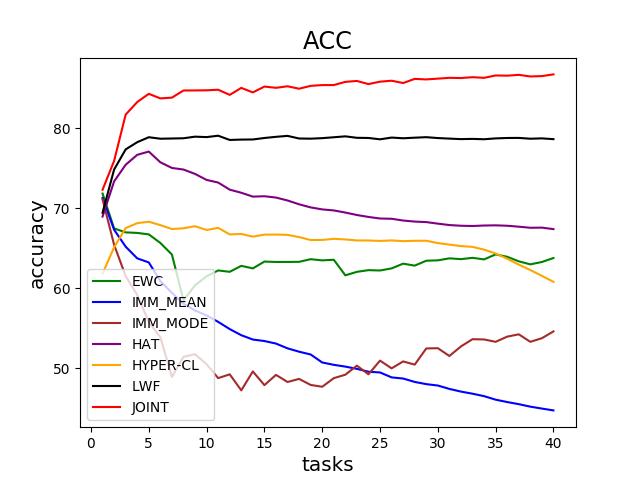} & 
  \includegraphics[width=0.42682436402\textwidth]{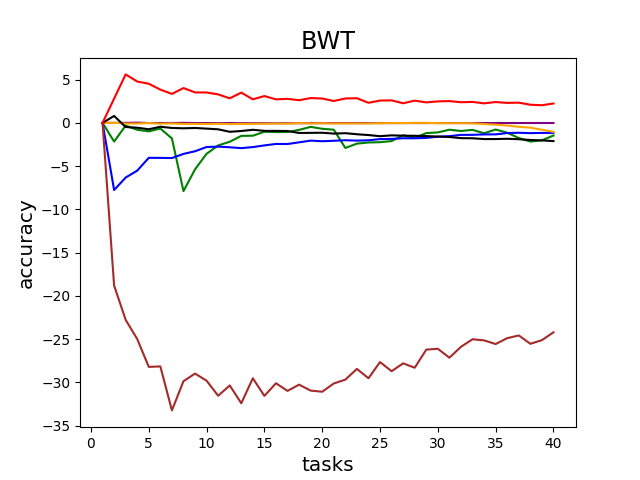}\\
(c) & (d)\\
  \end{tabular}
  \smallskip
  \caption{The evolution in time of the accuracy and the forgetting, for the best performing setting of each method average over 5 random seeds. $ACC$ (Eq.~\ref{eq:acc}) after learning task $t$ as a function of $t$. $BWT$ (Eq.~\ref{eq:bwt}) after learning task $t$ as a function of $t$.  (a) \& (b) $ACC$ \& $BWT$ results over time for CIFAR 20-Split and (c) \& (d) similar results over time for Tiny-ImageNet 40-Split.}
\label{fig:time}
\end{figure*}

The results are summarized in Tab.~\ref{table:aug_table}. As can be observed, applying augmentation in this setting leads to improvement in both ACC and BWT. Therefore, there is no trade-off between accuracy and forgetting. We emphasize that even though no augmentations protocol search was conducted and that the set of augmentations in use is rather small and simple, the performance boost is substantial.

\subsection{Comparison with other methods}
\label{sec:43}

We consider two regularization-based methods: EWC~\cite{kirkpatrick2017overcoming} and IMM~\cite{lee2017overcoming} and two parameter isolation methods: HAT~\cite{serra2018overcoming} and Hyper-CL~\cite{von2019continual}. ACL~\cite{Ebrahimi2020AdversarialCL} is considered as a recent hybrid method. As an upper bound for overall performance we consider a joint training method (JOINT), which for each incoming task, trains on the data of all tasks seen so far.
The hyper-parameters for EWC, IMM and HAT were the best found in~\cite{serra2018overcoming} and for Hyper-CL to the best found in~\cite{von2019continual}. For ACL, we quote the results mentioned in the paper, \ie for AlexNet-like architecture {with Dropout} (both private and shared) and no augmentations at all.

Following our findings for LwF, we opt to use all baseline methods with WRN-20-W5. However, we found that none of the baseline methods performs well with it. We found that some of the baseline methods are tightly coupled with the architecture originally presented in the paper. The authors of Hyper-CL~\cite{von2019continual} did an extensive hyperparameter search for both the hypernetwork and target architectures. They conclude that it is crucial to choose the right combination since it has a great effect on performance. Therefore, we used the best Hypernetwork-Target pair they found for the ``chunked'', more effective, version. This pair consists of a hypernetwork which has a linear layer that maps task and chunk embedding of size 32 each to a chunk of size 7000 of a ResNet-32 target network. Another coupling we found was for the HAT method, we could not achieve reasonable performance with an underlying ResNet architecture. We conjecture that the masking process in HAT needs to be adapted for usage with batch normalization layers, and report results with the AlexNet-like network presented by Serr{\`a} \etal \cite{serra2018overcoming}.

Both EWC and IMM, although not coupled with specific architecture, were found to be under-performing with WRN-20-W5, see appendix. We conjecture that the difference from LwF lies in the type of regularization term used by each method. LwF employs a `soft' regularization on the network output for previous tasks, which handles statistical shift due to batch normalization better than the weight-based regularization. For the comparison table we use the best evaluated architecture for each method.

All methods, except Hyper-CL and ACL, use separated fully-connected layers with a softmax output for each task as a final layer. Hyper-CL employs a separate generated network for each task, and ACL employs a separate 3-layer MLP with softmax output for each task on top of private and shared concatenation.

\noindent{\bf Training~} We made an effort to find the best training protocol for each method, based on the existing literature and initial experiments. 
For all methods except for Hyper-CL we followed the same training protocol described in Sec.~\ref{sec:net_effect}. For Hyper-CL, we use batch size 32 and with the Adam optimizer~\cite{kingma2014adam} with a learning rate of $0.001$. As for learning rate scheduling, Hyper-CL uses a validation accuracy to schedule the learning rate by dropping the learning rate with a factor of $(\sqrt{0.1})^{-1}$, if there is no improvement in the validation accuracy for 5 consecutive epochs. The Hyper-CL implementation further employs a custom multi-step scheduler adapted from Keras~\cite{chollet2015keras}. However, there is no early stopping in Hyper-CL. Also, no other regularization is used in any of the methods, except to the ones that are inherent to the method itself.

The Hyper-CL official implementation and the author's experiments use the test set for parameter selection in lieu of a proper validation set. We were able to fix and rerun the experiments in time only for the Hyper-CL experiments on CIFAR and not for the Hyper-CL experiments on Tiny-ImageNet. We observed that moving to an independent validation set reduces the performance of Hyper-CL on CIFAR by a significant margin. We, therefore,  view the results obtained for this method on Tiny-ImageNet as an upper bound for the method's performance. We note that (i) Hyper-CL is by far the slowest method out of all methods tested, and (ii) On Tiny-ImageNet even though the results of this method are positively biased, the method is not competitive.

The comparison to the literature methods is provided in Tab.~\ref{table:method_table} and summarized in Fig.~\ref{fig:best} for the best configuration for each method. 
Evidently, in contrast to the picture the literature paints, when a proper architecture and added augmentations are used, LwF, which is a simple regularization-base method, outperforms all other methods. The results also show that although IMM has evolved from EWC, both its variants are not competitive with EWC except for the smallest split (CIFAR 5-Split). When considering the augmentation mechanism, we have mixed results. Although augmentations increase ACC, they also increase forgetting for EWC and IMM-MEAN and only slightly reduce forgetting for IMM-MODE, which is still quite high. In contrast, for LwF, where we show that augmentations help to both ACC and BWT.

HAT as originally conceived (recall that it is not compatible with ResNets), has a very competitive ACC in CIFAR and even outperforms Hyper-CL for the longer and more challenging sequence of tasks from Tiny-ImageNet. It also further benefits from the augmentation. For Hyper-CL, we can see that although it has a smaller capacity (considering only the hypernetwork learnable parameters for capacity computation) it {outperforms all of the baselines for CIFAR when augmentation is used}. However, this advantage does not generalize to the Tiny-ImageNet dataset, {and it falls behind HAT, and even EWC}, for a longer sequence, which further emphasizes the need for comparison over a diverse set of experiments. To check if this shortcoming is a result of the capacity of the model, we experimented with larger models, both for the hypernetwork and target network.  We observed that the performance drops significantly in all experiments for the larger network. This result emphasizes the need for careful tuning of the Hyper-CL method, which is challenging since unlike other methods it requires the tuning of two architectures at once, which enlarges the space of possible hyper-parameters dramatically. We note also that \cite{von2019continual} reported that out of many architectures tried, the smallest ones showed the best performance-compression ratio.  

For ACL, we quote the results for {CIFAR 20-Split} with no augmentation from the paper itself \cite{Ebrahimi2020AdversarialCL}. The network used in the paper was similar to the one used by HAT. As the results show, ACL outperforms both HAT and Hyper-CL when no augmentation is used. LwF is not considered as a baseline in \cite{Ebrahimi2020AdversarialCL}. However, LwF outperforms ACL with WRN-20-W5 even without augmentation. We emphasize that the difference does not come from capacity, since both networks have a similar capacity as described in Tab.~\ref{table:arch_table}. 

We further analyze the performance by evaluating ACC and BWT after learning each task. Fig.~\ref{fig:time} shows the results for the longer sequences of tasks, 20 for CIFAR and 40 for Tiny-ImageNet (the results for the other experiments can be found in the appendix). One can observe that the methods differ in substantial ways. First, the non-LwF regularization methods, namely EWC and IMM, are not competitive with LwF since the early stages of the online training.  The results also indicate that although more careful tuning between the primary loss and the regularization loss could be made, there is a high degree of trade-off between forgetting and new learning in these methods. Where EWC and IMM-MEAN favor old tasks (low forgetting, low ACC) and IMM-MODE favors new tasks (high forgetting, comparable, or higher, final ACC to IMM-MEAN). Second, the same trade-off exists for HAT: while almost no forgetting exists, the accuracy for new tasks is lower. Since HAT is a parameter isolation method, we conjecture that it struggles to utilize the underlined architecture for learning new tasks. Third, while Hyper-CL and LwF seem close on CIFAR, an important difference is evident in Tiny-ImageNet. Looking at the profile of ACC for Tiny-ImageNet, Fig. \ref{fig:time} (c), shows that Hyper-CL struggles to learn new tasks after task 34 is learned, and the drop of accuracy is not due to forgetting, as is evident by the BWT plot in Fig.~\ref{fig:time} (d). Interestingly, this drop also enables EWC to outperform Hyper-CL through more consistent performance after the drop in task 8. Last, for LwF, in both CIFAR and Tiny-ImageNet, it enjoys the capability of learning new tasks and almost does not forget previous tasks. We conclude that, although LwF is a regularization based method, given the right architecture and augmentation, it can maintain both the ability to learn new tasks and to not forget old ones, even at the tails of long tasks sequence. 

{This emphasizes the need for a careful evaluation of each method. While EWC, IMM, HAT, and ACL outperform AlexNet-based LwF with Dropout architecture they fall short when dropout is removed and when selecting more appropriate architectures. The reason that these other methods do not suffer from Dropout is that they employ hard regularization on the weights which considers their importance. However, as Fig.~\ref{fig:time} shows, this type of regularization quickly results in a network utilization problem for fixed-size backbones. }

\section{Conclusions}

Many of the recent task-incremental publications~\cite{Li2019,serra2018overcoming,aljundi2018memory} compare with LwF and found their method to be superior. These conclusions seem to arise from the { little incentive authors have to explore the effect of the evaluation settings on prior work, or to invest effort in modernizing the form (\eg, architecture) of baseline methods.} However, LwF itself is built on top of solid knowledge-distillation foundations and, as we show, can be upgraded to become extremely competitive.

We demonstrate that the LwF method can benefit from a higher capacity (width-wise) and a network that employs residual connections as well as from augmentations. It is not obvious that the method would benefit from these changes, as many of the other methods cannot benefit from ResNets due to the challenges of applying batch normalization and the need to carefully control the capacity. Moreover, not all methods benefit from augmentations in both ACC and BWT.

Overall, our contributions are two-fold. First, we provide strong baselines for task-incremental methods, that form a solid foundation for comparing future methods. Second, we show the effect of added capacity, residual architectures, and regularization in the form of augmentation on {task-incremental} methods. Demonstrating sometimes paradoxical behavior, expected to improve performance but deteriorates it. We believe that LwF's ability to benefit from such improvements is a strong indication that this method would stand the test of time.

\section*{Acknowledgments}
 This project has received funding from the European Research Council (ERC) under the European Unions Horizon 2020 research and innovation programme (grant ERC CoG 725974).

{
\bibliographystyle{bib_style}
\bibliography{mbib}
}

\appendix
\section{ResNets architectures}

In section 4.1 of the main paper, we offered to use various ResNet architectures for LwF: RN-20, RN-32, RN-62, WRN-20-W2, and WRN-20-W5. All these networks share a common structure but differ in width or depth. This structure starts with a single convolutional layer of 16 filters with a kernel size of 3x3 and stride 1, followed by 3 groups of ``blocks''. Each group is parameterized by the number of blocks, width, and stride for the first block in the group. The baseline width (width factor equals 1) of each group is 16, 32, and 64, and strides 1, 2, and 2 respectively. 

To implement the blocks, the class of BasicBlock from the PyTorch framework is employed.  Each block contains 2 convolutional layers with a kernel size of 3x3 and a skip connection. The structure ends with an adaptive average pooling of size 1x1. Moreover, each convolutional layer is followed by a batch normalization layer and a ReLU activation function.

\noindent The parameters of the architectures in our work:
\begin{itemize}

\item{\bf RN-20} a width factor of 1 and 3 blocks in each group.

\item{\bf RN-32} a width factor of 1 and 5 blocks in each group.

\item{\bf RN-62} a width factor of 1 and 10 blocks in each group.

\item{\bf WRN-20-W2} a width factor of 2 and 3 blocks in each group.

\item{\bf WRN-20-W5} a width factor of 5 and 3 blocks in each group.
\end{itemize}

\section{LwF with AlexNet and data augmentations}

In the main text the best architecture is tested for LwF with data augmentations, namely WRN-20-W5. In this section we provide results for AlexNet-like architectures with augmentations as well, the results are provided in Tab.~\ref{table:lwf_alexnet_aug}. We observe that the data augmentations does not provide recovery from the harmful Dropout component in AlexNet-D. However, it does  provide performance boost for AlexNet-ND, as expected.

\begin{table*}[h]
\resizebox{\textwidth}{!}{
\centering
\begin{tabular}[t]{@{}l@{~}c@{~}c@{~}c@{~~}c@{~}c@{~~}c@{~}c@{~~}c@{~}c@{}}
\toprule
& & \multicolumn{2}{c}{CIFAR 5-Split}     & \multicolumn{2}{c}{CIFAR 10-Split}    & \multicolumn{2}{c}{CIFAR 20-Split}  & \multicolumn{2}{c}{Tiny-ImageNet 40-Split}    \\
\cmidrule(l{2pt}r{2pt}){3-4}
\cmidrule(l{2pt}r{1pt}){5-6}
\cmidrule(l{2pt}r{1pt}){7-8}
\cmidrule(l{2pt}r{1pt}){9-10}
Arch.  &  Aug.    & BWT            & ACC            & BWT            & ACC            & BWT            & ACC             & BWT            & ACC             \\
\midrule
AlexNet-D  & & $-39.9 \pm 1.4$  & $36.6 \pm 1.5$    & $-52.9 \pm 1.2$   & $28.1 \pm 1.3$   & $-54.4 \pm 1.1$   & $31.3 \pm 0.8$    & $-50.5 \pm 1.0$  & $25.0 \pm 0.4$   \\
AlexNet-D & \checkmark & $-46.2 \pm 1.8$ & $38.0 \pm 1.7$  & $-56.9 \pm 0.8$  &  $30.1 \pm 0.7$  & $-58.0 \pm 0.5$  & $31.6 \pm 0.3$ & $52.6 \pm 0.8$ & $25.9 \pm 0.5$    \\
AlexNet-ND  & & $-1.8 \pm 0.6$  & $56.6 \pm 1.1$    & $-2.9 \pm 0.2$   & $67.0 \pm 1.0$   & $-3.1 \pm 0.3$   & $75.5 \pm 0.6$    & $-2.8 \pm 0.3$  & $66.9 \pm 0.8$ \\
AlexNet-ND &  \checkmark & $-0.5 \pm 0.4$ & $69.5 \pm 1.1$ &   $-0.7 \pm 0.3$ & $76.7 \pm 0.9$ & $-0.9 \pm 0.2$ & $83.5 \pm 0.5$ & $-1.4 \pm 0.3$   & $73.2 \pm 0.7$  \\
\end{tabular}}
    \caption{LwF results with AlexNet-like architecture with data augmentations. all results are produced by us and are averaged over five runs with standard deviations. D=Dropout, ND=No Dropout. }
    \label{table:lwf_alexnet_aug}
\end{table*}

\begin{figure*}[t]
  \centering
    \begin{tabular}{@{}c@{~}c@{~}}
  \includegraphics[width=0.5\textwidth]{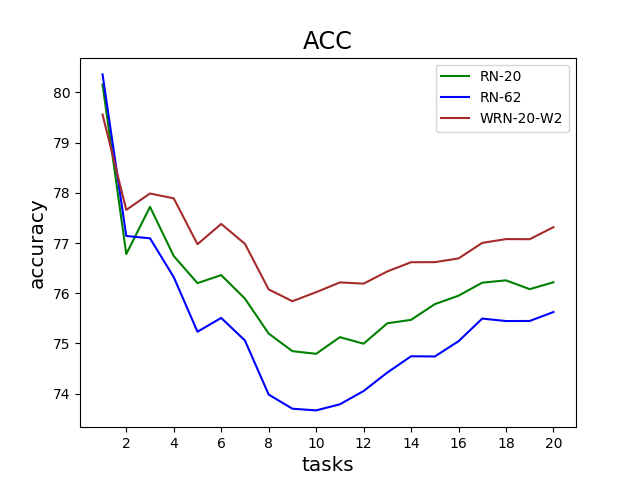} & \includegraphics[width=0.5\textwidth]{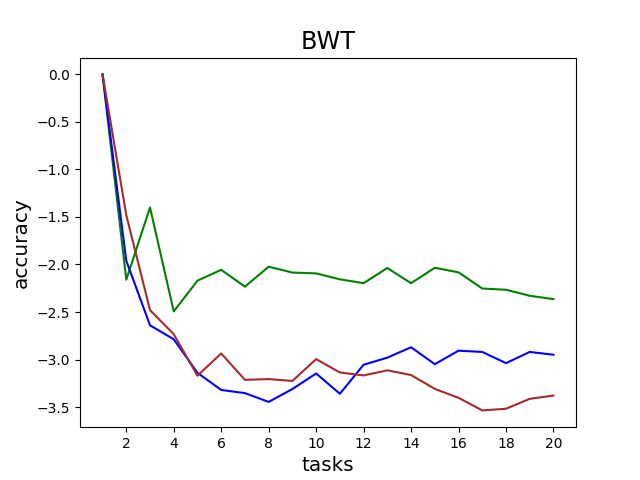} \\
(a) & (b)
  \end{tabular}
  \caption{\small The evolution in time of the accuracy and the forgetting for CIFAR 20-Split with LwF and different width and depth architectures, average over 5 random seeds. No augmentation used in these experiments. (a) $ACC$ (Eq. 1) after learning task $t$ as a function of $t$. (b) $BWT$ (Eq. 2) after learning task $t$ as function of $t$. }
\label{fig:width_vs_depth}
\end{figure*}

\section{Width vs. depth for LwF}

In Fig.~\ref{fig:width_vs_depth} we offer another view on the effect of different depth and width for LwF. The results are provided for the baseline ResNet architecture, RN-20, and two comparable capacity architectures. One with greater depth, RN-62, and another with greater width, WRN-20-W2. The results show that although RN-62 and WRN-20-W2 share a similar amount of forgetting, from task 2 onward RN-62 under-performs with respect to ACC. 

This suggests that LwF with a deeper ResNet network is struggling to acquire new knowledge while keeping the previous one. Comparing RN-62 with RN-20 highlights a more severe problem where LwF is struggling to utilize deeper networks both in terms of ACC and BWT. However, increased width has a positive effect on performance over time, even at the price of increased forgetting. Fortunately, we were able to mitigate this increased forgetting with data augmentations, which not only reduced forgetting substantially but also increased ACC.

{
\section{EWC and IMM with WRN-20-W5}

In our experiments we found EWC and IMM (both MEAN and MODE variants) to perform poorly with ResNet architectures and specifically with WRN-20-W5. The results, for this architecture,  can be found in Tab.~\ref{table:ewc_and_imm_resnet}. As can be seen, using WRN-20-W5 the methods are not competitive and perform lower than when using the AlexNet-like architecture, as quoted in the main paper. This performance gap suggests that the methods require modifications in order to enjoy more modern architecture, like ResNet. We attribute this to the challenge imposed by the batch normalization layers.}

\begin{table*}[t]
\resizebox{\textwidth}{!}{
\centering
\begin{tabular}[t]{@{}l@{~}c@{~}c@{~}c@{~~}c@{~}c@{~~}c@{~}c@{~~}c@{~}c@{}}
\toprule
& & \multicolumn{2}{c}{CIFAR 5-Split}     & \multicolumn{2}{c}{CIFAR 10-Split}    & \multicolumn{2}{c}{CIFAR 20-Split}  & \multicolumn{2}{c}{Tiny-ImageNet 40-Split}    \\
\cmidrule(l{2pt}r{2pt}){3-4}
\cmidrule(l{2pt}r{1pt}){5-6}
\cmidrule(l{2pt}r{1pt}){7-8}
\cmidrule(l{2pt}r{1pt}){9-10}
Method &  Aug.    & BWT            & ACC            & BWT            & ACC            & BWT            & ACC             & BWT            & ACC             \\
\midrule
EWC & & $-11.0 \pm 2.4$ & $46.8 \pm 2.1$  & $-24.8 \pm 3.6$  &  $39.8 \pm 2.6$  & $-33.5 \pm 5.5$  & $40.9 \pm 5.3$ & $-31.4 \pm 2.0$ & $34.8 \pm 1.6$    \\
EWC &  \checkmark & $-11.6 \pm 3.9$ & $60.1 \pm 4.4$ &   $-31.9 \pm 2.6$ & $46.8 \pm 2.4$ & $-45.7 \pm 4.1$ & $38.2 \pm 3.4$ & $-45.1 \pm 3.1$   & $31.1 \pm 3.5$  \\
IMM-MEAN &   & $-12.3 \pm 8.5$ & $24.6 \pm 8.7$  & $-3.5 \pm 5.6$ & $27.3 \pm 4.4$  & $-2.9 \pm 1.3$ & $33.3 \pm 2.0$ & $+0.2 \pm 1.5$   & $28.1 \pm 1.3$   \\
IMM-MEAN & \checkmark & $-16.9 \pm 4.7$ & $29.3 \pm 3.2$ & $-4.9 \pm 2.5$  & $29.4 \pm 3.1$  & $-3.3 \pm 2.1$ & $30.9 \pm 1.3$ & $-1.6 \pm 4.0$ &  $26.8 \pm 3.0$   \\
IMM-MODE &   & $-22.7 \pm 6.3$ & $39.4 \pm 3.9$   &  $-34.8 \pm 4.0$  & $34.5 \pm 3.1$ & $-47.3 \pm 4.0$  &  $30.3 \pm 3.3$  & $-42.5 \pm 2.1$ & $27.5 \pm 1.4$ \\
IMM-MODE &  \checkmark & $-39.8 \pm 2.1$ & $44.0 \pm 2.1$ & $-52.0 \pm 3.3$ &  $35.2 \pm 2.7$  & $-58.8 \pm 5.4$  & $30.2 \pm 5.2$ & $-52.4 \pm 2.7$ & $26.4 \pm 2.5$  \\
\end{tabular}}
    \caption{EWC and IMM results with WRN-20-W5. all results are produced by us and are averaged over five runs with standard deviations.}
    \label{table:ewc_and_imm_resnet}
\end{table*}

\section{ACC and BWT over time}

\begin{figure*}[h]
  \centering
    \begin{tabular}{@{}c@{~}c@{}}
  \includegraphics[width=0.485\textwidth]{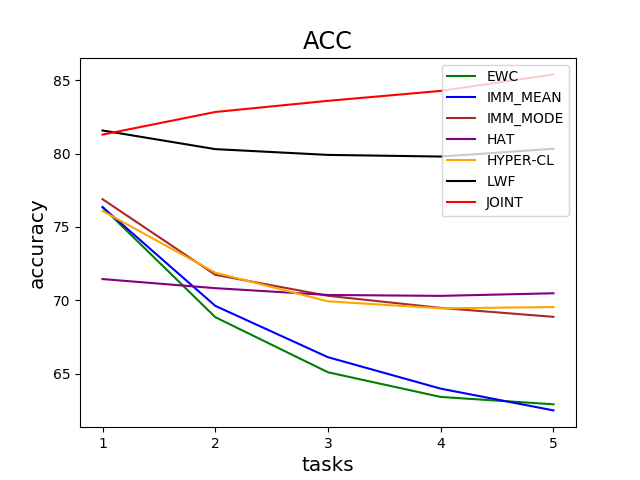} & \includegraphics[width=0.485\textwidth]{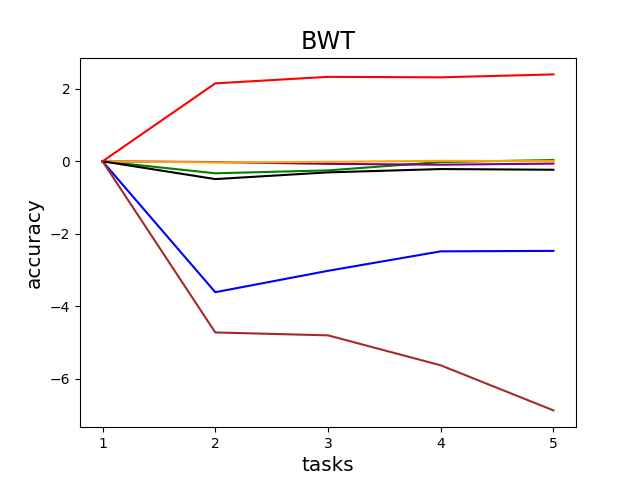} \\
  (a) & (b) \\
  \includegraphics[width=0.485\textwidth]{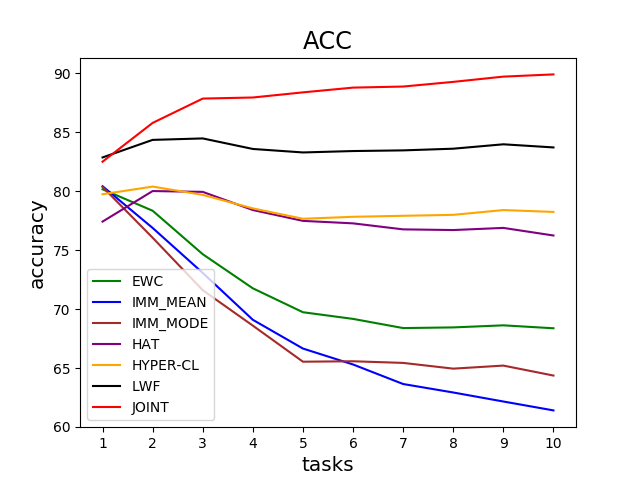} & 
  \includegraphics[width=0.485\textwidth]{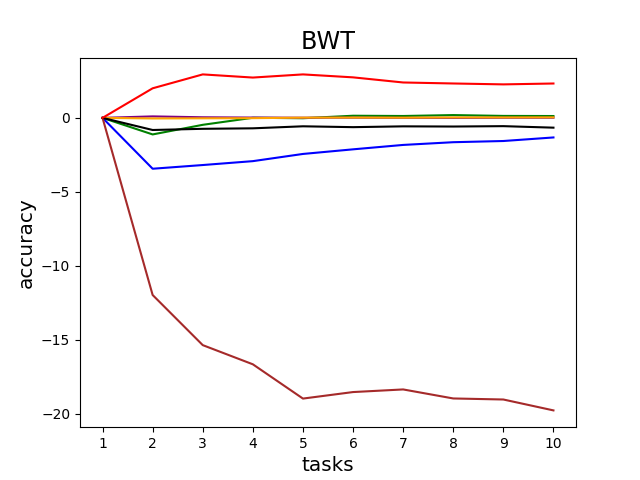}\\
  (c) & (d)\\
  \end{tabular}
  \smallskip\caption{\small The evolution in time of the accuracy and the forgetting, for the best performing setting of each method average over 5 random seeds. $ACC$ (Eq. 1) after learning task $t$ as a function of $t$. $BWT$ (Eq. 2) after learning task $t$ as function of $t$.  (a) \& (b) results over time for CIFAR 5-Split and (c) \& (d) results over time for CIFAR 10-Split.}
\label{fig:time_sup}
\end{figure*}

In Fig.~\ref{fig:time_sup} we provide the BWT and ACC scores after learning each task for CIFAR-100 with 5 and 10 splits. These results were omitted from the main text for brevity and provided here as complementary results.

Similarly to the results shown in the paper (main text Fig. 2), the advantage of LwF over the baseline methods is evident. LwF can learn new tasks with a similar level of performance to the previous ones while maintaining the knowledge from the previous tasks. In contrast, both EWC and IMM fail to do so. For HAT, the difference in performance between different CIFAR-100 splits, {where the performance is more stable for a short sequence of tasks,} could point to an insufficient per task capacity problem. However, since LwF can both learn new tasks and maintain old ones with similar capacity, this points to an under-utilization of the network capacity. Thus, we suspect that HAT is not scalable for long task sequences even with larger networks. Although HyperCL seems to have very competitive results for these splits, its shortcoming is revealed in the main paper, looking at a longer sequence of tasks, such as Tiny-ImageNet.

\end{document}